\documentclass[10pt,twocolumn,letterpaper]{article}

\usepackage[pagenumbers]{cvpr} 

\usepackage[ruled,vlined]{algorithm2e}
\usepackage{multirow}

\definecolor{cvprblue}{rgb}{0.21,0.49,0.74}
\usepackage[pagebackref,breaklinks,colorlinks,allcolors=cvprblue]{hyperref}
\usepackage{outlines}

\def\paperID{6035} 
\def\confName{CVPR}
\def\confYear{2025}

\title{Consistency-aware Self-Training for Iterative-based Stereo Matching}

\author{First Author\\
Institution1\\
Institution1 address\\
{\tt\small firstauthor@i1.org}
\and
Second Author\\
Institution2\\
First line of institution2 address\\
{\tt\small secondauthor@i2.org}
}
\author{
    Jingyi Zhou$^{1}$\thanks{Equal Contribution},
    Peng Ye$^{1,3,4*}$,
    Haoyu Zhang$^{1*}$,
    Jiakang Yuan$^{1}$\\
     Rao Qiang$^{2}$,
     Liu YangChenXu$^{2}$,
     Wu Cailin$^{2}$,
     Feng Xu$^{1}$,
    Tao Chen$^{1}$\thanks{Corresponding author.} 
    \\
    {\normalsize $^{1}$Fudan University}
    {\normalsize $^{2}$Xiaomi Inc} 
    {\normalsize $^{3}$Shanghai AI Laboratory }
    {\normalsize $^{4}$The Chinese University of Hong Kong}
}

\begin{document}
\maketitle
\begin{abstract}
Iterative-based methods have become mainstream in stereo matching due to their high performance. However, these methods heavily rely on labeled data and face challenges with unlabeled real-world data. 
To this end, we propose a consistency-aware self-training framework for iterative-based stereo matching for the first time, leveraging real-world unlabeled data in a teacher-student manner. We first observe that regions with larger errors tend to exhibit more pronounced oscillation characteristics during model prediction.
Based on this, we introduce a novel consistency-aware soft filtering module to evaluate the reliability of teacher-predicted pseudo-labels, which consists of a multi-resolution prediction consistency filter and an iterative prediction consistency filter to assess the prediction fluctuations of multiple resolutions and iterative optimization respectively. Further, we introduce a consistency-aware soft-weighted loss to adjust the weight of pseudo-labels accordingly, relieving the error accumulation and performance degradation problem due to incorrect pseudo-labels. 
Extensive experiments demonstrate that our method can improve the performance of various iterative-based stereo matching approaches in various scenarios. In particular, our method can achieve further enhancements over the current SOTA methods on several benchmark datasets.
\end{abstract}    
\section{Introduction}
\label{sec:intro}

\begin{figure}[t]
\centering
\includegraphics[width=0.98\columnwidth]{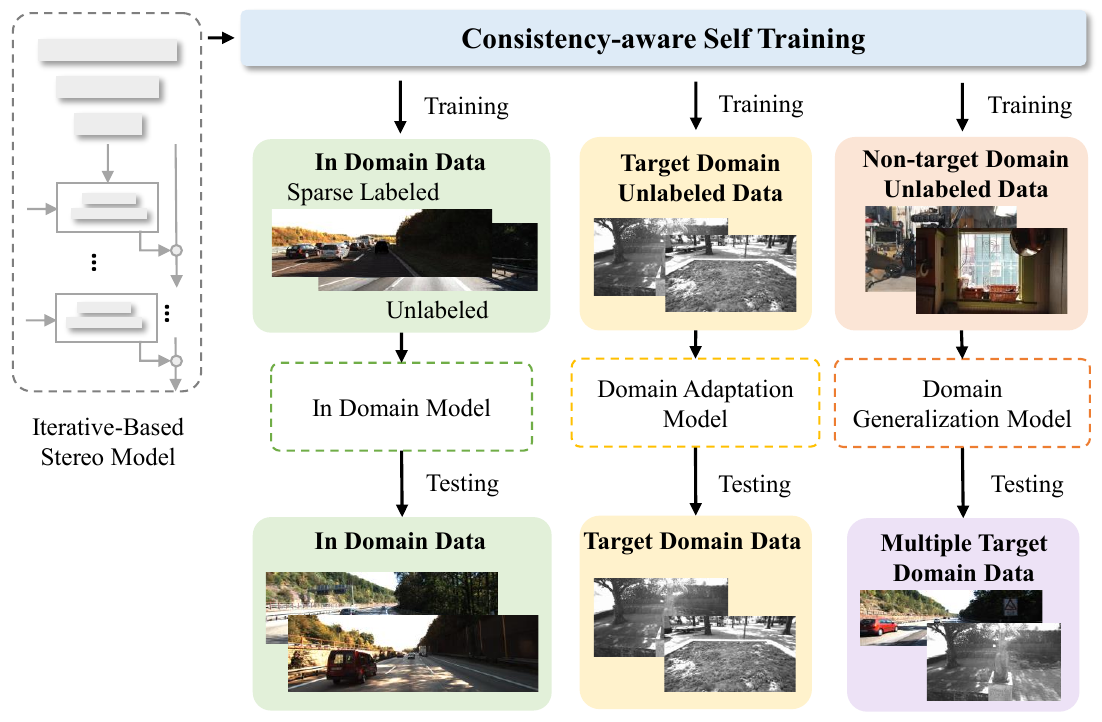} 
\vspace{-3mm}
\caption{Illustration of our Consistency-aware Self-Training (CST) method, which can leverage unlabeled data in multiple scenarios to boost the iterative-based stereo-matching model.}
\label{fig1}
\vspace{-4mm}
\end{figure}

Stereo matching aims to predict the pixel-wise disparity between a pair of rectified images~\cite{988771}. With the rapid development of 3D vision tasks and their applications such as robotics and autonomous driving, stereo matching has become a fundamental vision task to provide depth information in the real 3D world. Although existing stereo matching methods have achieved remarkable performance on certain benchmark datasets~\cite{crestereo,wang2024selective}, they heavily rely on the number and quality of labeled data. However, due to hardware and environmental constraints, obtaining high-quality labels for stereo data is known to be costly and sometimes even infeasible. Thus, current labeled stereo data is predominantly composed of synthetic datasets, leading to suboptimal performance when applied to unlabeled real-world data. Unlike the scarcity of labeled real-world stereo data, unlabeled data is relatively easier to acquire while often ignored by previous works.

Self-training has been proven to be an effective approach to leveraging unlabeled data to enhance model performance and generalization ability via a teacher-student manner. As a result, a few recent works~\cite{Yue_2024_CVPR,Xu2024AUA} attempt to apply self-training on cost-volume based stereo matching methods, where the predicted disparities are based on the cost-volume matrix and the reliability of teacher-predicted pseudo-labels is obtained via the matching probability distribution of cost volume.
However, these cost-volume-based self-training methods still encounter the following challenges: 1) As the iterative-based methods led by RAFT-Stereo~\cite{raft-stereo} gradually become mainstream, these methods may fail to adapt effectively since the iterative-based methods employ recurrent units to update the disparity with multi-source information and sometimes without calculating the whole cost volume. 2) The reliability computation of these methods resembles classification, treating the pixels along the epipolar line as independent discrete labels. It focuses on single-point concentration while neglecting the spatial concentration along the epipolar axis, which may lead to a significant discrepancy between the predicted and true reliability. 3) These methods follow the all-or-nothing threshold filtering approach, where pseudo-labels are either selected or discarded with no intermediate state. Considering that pseudo-labels within the threshold are not equally reliable and those not selected still hold potential training value, such a rigid binary selection approach poses a bottleneck for performance improvement, particularly in challenging areas.

To address these issues, we propose a novel consistency-aware self-training framework for iterative-based stereo matching for the first time. Our framework is based on the insightful observation that regions with larger errors tend to exhibit more pronounced oscillation characteristics during model prediction. 
Motivated by this, we propose a consistency-aware soft filtering module that evaluates the reliability of teacher-predicted pseudo-labels for soft filtering unreliable pseudo-labels to mitigate error accumulation and performance degradation.
It consists of a multi-resolution prediction consistency filter and an iterative prediction consistency filter. The former is used to assess the prediction fluctuations of multiple resolutions since error-prone pixels often display instability and inconsistency with the changing of image resolution. The latter is used to assess the prediction fluctuations of iterative optimization since error-prone areas strongly correlate with fluctuating regions of iterative predictions.
Further, we introduce a consistency-aware soft-weighted loss to assign higher weights to predictions with greater reliability, reducing the impact of teacher-predicted incorrect pseudo-labels while keeping the student's possibility to learn from diverse and challenging areas.
Our method: 1) introduces an effective self-training strategy for iterative-based stereo matching without relying on the cost volume; 2) computes the reliability suitable for regression by accounting for the concentration along the epipolar axis rather than isolated peak values, which is more accurate. 3) employs soft filtering to avoid absolute binary selection, ensuring a more nuanced pseudo-label selection process.

To demonstrate the effectiveness of the proposed method, comprehensive experiments are conducted across multiple challenging scenarios, including in domain, domain adaptation and domain generalization, as shown in Fig.~\ref{fig1}. The experimental results verify that our method significantly enhances the performance and generalization capability of various iterative-based stereo matching models, enabling them to perform well not only on unlabeled datasets but also on real-world datasets that have not been encountered during training. Especially, our method can achieve further improvements over state-of-the-art methods and
on the Middlebury, KITTI2015, and ETH3D datasets, 

Our contributions are briefly summarized as follows:
\begin{itemize}
    \item We propose a novel consistency-aware self-training framework for iterative-based stereo matching for the first time,
    based on the insightful observation that regions with larger errors tend to exhibit more pronounced oscillation characteristics during model prediction.
    \item We
    design a consistency-aware soft filtering module to assess the pixel-level reliability of teacher-predicted pseudo-labels from spatial and temporal dimensions, which includes a multi-resolution prediction consistency filter and an iterative prediction consistency filter. 
    \item We introduce a consistency-aware soft-weighted loss for the training of the student model, which reduces the impact of the teacher model's errors on the student model while maintaining the student model's possibility to learn from diverse and challenging areas.
    \item Our method achieves impressive results in various scenarios, including in domain, domain adaptive and domain generalization, achieving new SOTA results among the published works. 
\end{itemize}


\section{Related Work}
\subsection{Stereo Matching}
\textbf{Cost-volume-based methods.} Since ~\citeauthor{zbontar2015computing} first applied machine learning methods to the feature matching stage of stereo matching, traditional matching algorithms~\cite{boykov2001fast,klaus2006segment} have gradually replaced by learning-based methods~\cite{guo2019group,9879745,10330699}. Initially, cost-volume-based methods led the development. PSMNet~\cite{psmnet} utilized feature concatenation to build a 4D cost volume. HITNet ~\cite{tankovich2021hitnet} introduced a swift multi-resolution initialization phase, differentiable 2D geometric propagation, and warping mechanisms that balance speed and accuracy. Although these methods have made great progress compared to traditional methods, the huge memory consumption of these methods for inferring high-resolution images limits their further improvement. 

\textbf{Iterative-based methods.} In recent years, more and more works focused on iterative-based stereo matching. RAFT-Stereo~\cite{raft-stereo} firstly explored the use of iterative multi-scale update blocks composed of GRUs~\cite{chung2014empirical} to produce the final disparity map by progressing from coarse to fine. CREStereo~\cite{crestereo} employed an adaptive group correlation layer to mitigate the effects of cameras not being perfectly aligned in reality. IGEV-Stereo~\cite{igevstereo} introduced an additional Geometry Encoding Volume to compensate for the missing non-local geometry information in all-pairs correlations. 
EAI-Stereo~\cite{zhao2022eai} and DLNR~\cite{zhao2023high} explored the use of LSTM~\cite{hochreiter1997long} in iterative module. Selective-Stereo~\cite{wang2024selective} proposes a Selective Recurrent Unit to relieve the loss of detailed information. 
Even though these iterative-based methods performed well in public benchmarks, they still face challenges in real-world scenes due to the scarcity of labeled data. This drives us to think about how to leverage the unlabeled data to improve not only the benchmark performance but also the generalization ability of these models.

\subsection{Self Training}
Self-training is a classical method~\cite{Lee2013PseudoLabelT} and has been widely used in various fields, such as image classification~\cite{CascanteBonilla2020CurriculumLR}, object detection~\cite{Li2020ImprovingOD}, and semantic segmentation~\cite{9880151}. 
The typical approach to self-training involves using a teacher-student framework with pseudo-labeling to leverage unlabeled data during training. However, self-training often encounters confirmation bias, where the student inherits and reinforces the teacher's errors. To mitigate this, two strategies are commonly employed. First, applying strong perturbations to the student's input can help decouple predictions~\cite{9880151,Yang2024DepthAU}. Second, filtering pseudo-labels enhances their reliability, preventing the propagation of errors from teacher to student.
In stereo matching, self-training has been applied to cost-volume-based methods~\cite{Yue_2024_CVPR,Xu2024AUA}, which use cost-volume to assess label reliability. However, these methods are not effectively applicable to iterative-based approaches that predict prediction maps with recurrent networks receiving multi-source information. This research gap encourages us to explore self-training methods applicable to iterative-based approaches.
\begin{figure*}[t]
\centering
\includegraphics[width=0.92\textwidth]{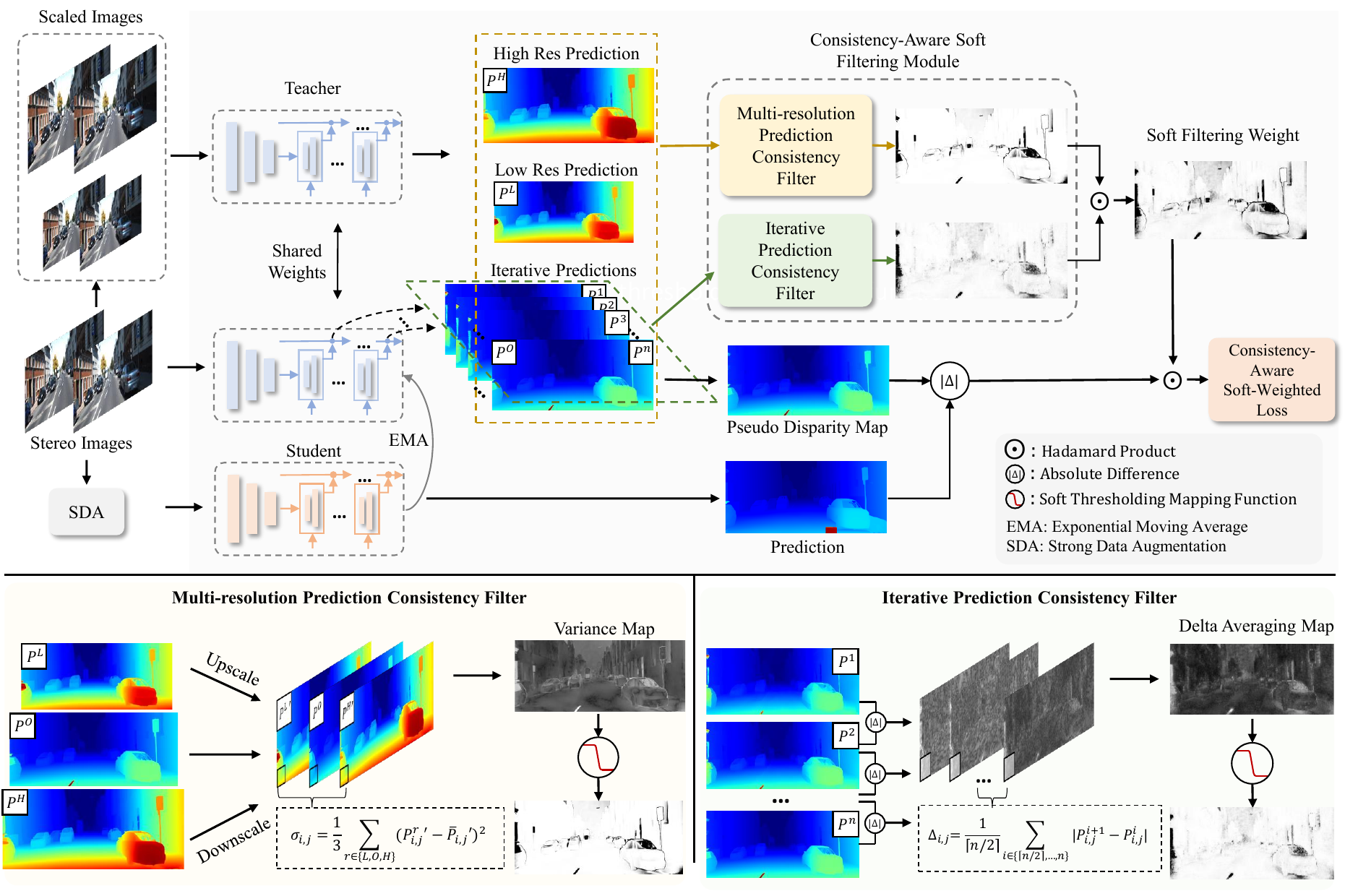} 
\vspace{-3mm}
\caption{Overview of our proposed CST-Stereo. Our framework leverages unlabeled stereo data in a teacher-student manner. In detail, the student receives strongly augmented images and learns from the teacher's predictions, with model parameters updated in a delayed manner with EMA for cyclic enhancement. Further, the consistency-aware soft filtering module is applied to evaluate the reliability of teacher-predicted pseudo-labels, which includes a multi-resolution prediction consistency filter and an iterative prediction consistency filter.
Finally, the consistency-aware soft-weighted loss is calculated for optimization.}
\label{fig2}
\vspace{-4mm}
\end{figure*}

\section{Method}

The overview of CST-Stereo is shown in Fig.~\ref{fig2}, given an unlabeled stereo pair, we first obtain the pseudo-labels using the teacher model. Then, the multi-resolution consistency score $w_{rc}$ and iterative consistency score $w_{ic}$ can be obtained by the consistency-aware soft filtering module (CSF) which is composed of a multi-resolution prediction consistency filter (MRPCF) and an iterative prediction consistency filter (IPCF). CSF assesses the reliability of the pseudo-labels for soft filtering unreliable pseudo-labels to mitigate error accumulation caused by the inaccurate predictions of the teacher model. Besides, the final soft filtering weights can be calculated according to $w_{rc}$ and $w_{ic}$ and can be utilized as the weight of pseudo-label supervision for the student model to update the student model. Finally, the teacher model is updated using exponential moving average (EMA) after several training iterations, inspired by the DINO~\cite{caron2021emerging}. Below we first give the problem definition and then detail each module of our design.

\begin{algorithm}[t]
\caption{Workflow of CST-Stereo}
\label{alg:algorithm1}
\KwIn{Labeled dataset $\mathcal{D}^l$, Unlabeled dataset $\mathcal{D}^u$, Strong augmentations $\mathcal{A}^s$, Upscale $\mathcal{F}^u$, Downscale $\mathcal{F}^d$, Teacher/student model $T/S$, The update interval $K$ of the teacher model}
\KwOut{Self-trained student model $S$}
Train a model on $\mathcal{D}^l$ with weight $\mathcal{W}$.

Initialize $T$ and $S$ with $\mathcal{W}$ and set current iteration $k=0$.



\For {$X \subset \mathcal{D}^u$} {
$X^H,X^L = \mathcal{F}^u(X),\mathcal{F}^d(X)$

$P^H,P^O,P^L = T(X^H),T(X),T(X^L)$

Intermediate predictions of $T(X)$: $\{P^i\}^{n}_{i=\left \lceil n/2 \right \rceil}$.

Calculate $w_{rc}$ and $w_{ic}$ using Eq.~(\ref{eq2}) and Eq.~(\ref{eq4}) based on  $P^H,P^O,P^L$, and $\{P^i\}^{n}_{i=\left \lceil n/2 \right \rceil}$.


Calculate $w_{soft}$ using Eq.~(\ref{eq5}) based on $w_{rc}$, $w_{ic}$.

Calculate the prediction of student model $\hat{P}=S(\mathcal{A}^s(X))$.

Update $S$ to minimize $L_{ST}$ calculated by Eq.~(\ref{eq6}) based on $w_{soft}$, $\hat{P}$, and $P^O$.

$k \leftarrow k+1$

\eIf {$k \% K == 0$} {
        $\theta_T \leftarrow \lambda \theta_T + (1-\lambda) \theta_S$
    }{pass}
}

\Return{$S$}
\end{algorithm}

\subsection{Problem Definition}
The workflow of our framework is illustrated in Algorithm~\ref{alg:algorithm1}. Given a labeled dataset $\mathcal{D}^l$ and an unlabeled dataset $\mathcal{D}^u$ (\textit{e.g.}, labeled synthetic data and unlabeled real-world data), we first train a model on $\mathcal{D}^l$ to obtain the initial weights for both the teacher model $T$ and the student model $S$. Then, given an unlabeled stereo pair $X = (I_l, I_r)$ sampled from $\mathcal{D}^u$, our goal is to improve the performance of iterative-based stereo matching models by leveraging unlabeled data in a self-training manner.

\subsection{Consistency-Aware Soft Filtering Module}

Self-training has been proven to effectively utilize unlabeled data to improve the model performance and generalization ability. However, the effectiveness of self-training heavily relies on the quality of the pseudo-labels. Since pseudo-labels often contain inaccurate predictions, directly training on the pseudo-labels without filtering may result in poor model performance as shown in Tab.~\ref{tab:ab1}.
To obtain more accurate pseudo-labels and avoid performance degradation caused by noisy supervision, we propose a consistency-aware soft filtering module (CSF) that filters pseudo-labels based on the estimated reliability. Instead of employing the commonly used binary filtering strategy, we adopt a soft thresholding approach by assigning each pixel an independent weight according to calculated reliability. 
This design reduces the potential impact of errors, thereby ensuring the stability of model training while still allowing the model to learn from challenging regions. Otherwise, some difficult areas such as edges and occluded regions might never be adequately learned.

\subsubsection{Multi-resolution Prediction Consistency Filter}
\begin{figure}[t]
\centering
\includegraphics[width=0.88\columnwidth]{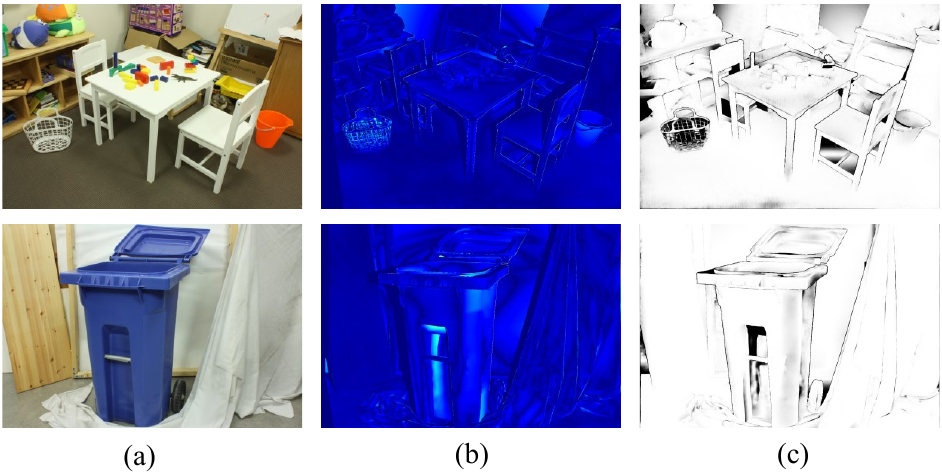} 
\vspace{-3mm}
\caption{Correlations between error regions and multi-resolution prediction consistency. (a) Referenced image. (b) Error map. (c) Multi-resolution prediction consistency map (Darker areas denote lower consistency).}
\label{fig3}
\vspace{-3mm}
\end{figure}

To filter noisy prediction with low reliability, we first develop a multi-resolution prediction consistency filter (MRPCF) based on our observation. 
\textbf{Observations:} As a pixel-level dense disparity regression task, stereo matching is particularly sensitive to changes in spatial scale information. Generally, at lower resolutions, the model demonstrates better global semantic capture and is less affected by noise, while at higher resolutions, it focuses more on detailed information. Due to the transitions, the predictions for some pixels become inconsistent. According to the experiments and analysis, we identify a significant correlation between the consistency of pixel predictions across different resolutions and their reliability, as shown in Fig.~\ref{fig3}. Specifically, pixels that maintain consistent predictions across varying resolutions are more likely to be reliable. This discovery provides a robust mechanism for identifying and addressing areas of uncertainty within the model's output. 

\textbf{Design:} Based on the observation, we first evaluate the accuracy of the pseudo disparity map by calculating the consistency of the multi-resolution prediction. Specifically,  we calculate the pixel-wise variance among predictions from the high, original, and low resolutions. In detail, the teacher model takes upscaled, original, and downscaled stereo images: $X^H$, $X$, and $X^L$ as inputs to generate disparity predictions: $P^H$, $P^O$, and $P^L$ respectively. The variance map $\sigma$ is then calculated as:
\begin{equation}
\label{eq1}
\sigma_{i,j} = \frac{1}{3} \sum_{r \in \{H,O,L\}} ({P_{i,j}^{r}}^{\prime}-\bar{P}_{i,j}^{\prime})^2,
\end{equation}
where ${P_{i,j}^{r}}^{\prime}$ represents the resized disparity map of $P_{i,j}^{r}$ and $\bar{P}^{\prime}$ is the average of ${P^H}^{\prime}$, ${P^O}^{\prime}$, and ${P^L}^{\prime}$. Pixels with lower variance indicate that their predicted values are more consistent across spatial transformations without significant deviations, and vice versa. Subsequently, a soft thresholding mapping function is employed to derive the reliability weight for each pixel, converting the variance into a reliability score within the 0-1 range. The function can be written as follows: 
\begin{equation}
\label{eq2}
w_{rc} = \frac{1}{1+e^{-\varepsilon_1(\sigma - \tau_1)}},
\end{equation}
where $\tau_1$ denotes the soft threshold and $\varepsilon_1$ is a scale factor. In this scheme, the greater the variance, the lower the reliability weight assigned to that pixel, with weights approaching zero for pixels with excessively high variance. The mapping function not only assigns varying weights to pixels based on their reliability but also ensures that pixels with excessive errors have minimal impact on the training process, providing a robust mechanism to weigh pixel predictions based on multi-resolution prediction consistency.

\subsubsection{Iterative Prediction Consistency Filter}
\begin{figure}[t]
\centering
\includegraphics[width=0.88\columnwidth]{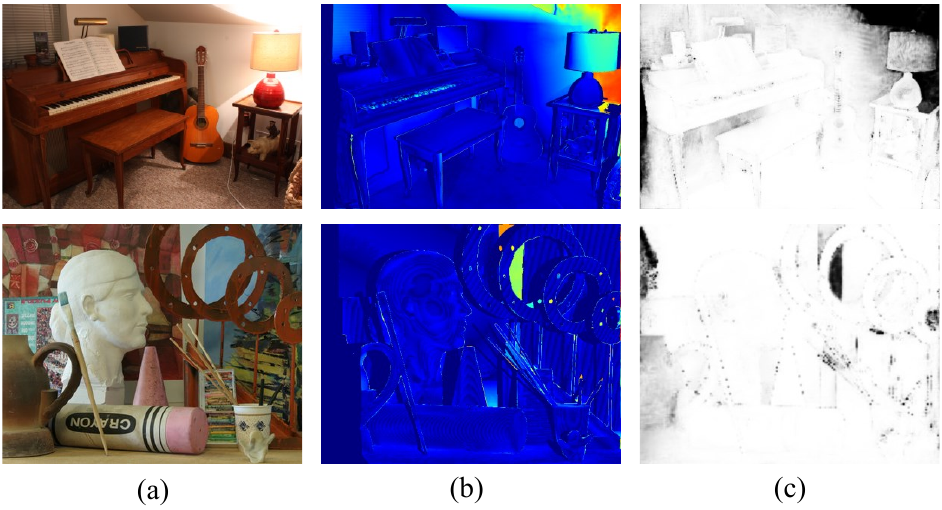} 
\vspace{-3mm}
\caption{Correlations between error regions and iterative prediction consistency. (a) Referenced image. (b) Error map. (c) Iterative prediction consistency map (Darker areas denote lower consistency).}
\label{fig4}
\vspace{-3mm}
\end{figure}

Although MRPCF performs well for resolution-sensitive errors, which often occur at edges and discontinuous regions, it still falls short in handling complex scenarios and ambiguous areas. Thus, we propose a iterative prediction consistency filter (IPCF) to complement the spatial perspective.
\textbf{Observations:} Iterative-based stereo matching model uses a recurrent neural network, often composed of GRUs, to incrementally refine predictions from either scratch or a rough estimate, aiming to approximate the true disparity values. During this process, the iterative update module integrates multi-source information, such as local cost values, contextual feature, and geometric information, to produce updated predictions. Generally, predictions initially undergo significant changes but gradually converge to a relatively stable value. However, we observed that some pixels exhibit oscillatory behavior in the later stages of iteration, which strongly correlates with error-prone regions, as shown in Fig.~\ref{fig4}. The oscillations likely indicate inconsistencies among the multi-source information in those areas, preventing the model from confidently converging to a single value. Such results also provide us with valuable insights into assessing the reliability of the model predictions.

\textbf{Design:} According to the observation, we propose an iterative prediction consistency filter that aims at finding inaccurate areas in the pseudo label through unstable disparity prediction in iterations. 
Specifically, we calculate the average of the difference map between the adjacent iteration rounds as follows:
\begin{equation}
\label{eq3}
\Delta_{i,j} = \frac{1}{\left \lceil n/2 \right \rceil} \sum_{k \in \{ \left \lceil n/2 \right \rceil , \ldots ,n\}} |P_{i,j}^{k+1}-P^k_{i,j}|,
\end{equation}
where $n$ is the total number of iterations, $P^k$ is the prediction of the $k$-th iteration. $\Delta_{i,j}$ can serve as an indicator of the prediction's stability over iterations. A smaller $\Delta_{i,j}$ implies that the prediction of the current pixel has stabilized after balancing the diverse information sources and thus is more reliable. Similar to the former section, a soft thresholding mapping function is used to shift the delta averaging map to reliability weight:
\begin{equation}
\label{eq4}
w_{ic} = \frac{1}{1+e^{-\varepsilon_2(\Delta - \tau_2)}},
\end{equation}
where $\tau_2$ and $\varepsilon_2$ represent the soft threshold and scale factor, respectively.

\begin{table*}[t]
    \centering
    \scalebox{0.85}{
    \begin{tabular}{l|ccc|cccc|ccc}
     \toprule
     \multirow{2}{*}{Method}  & \multicolumn{3}{c|}{ KITTI 2015} &\multicolumn{4}{c|}{Middlebury} &\multicolumn{3}{c}{ETH3D}\\
      & D1-bg & D1-fg & D1-all &  bad 1.0 &bad 2.0 & bad 4.0 &Avgerr & bad 1.0 & bad 2.0 & Avgerr\\
     \midrule
    PSMNet~\cite{psmnet}   & 1.86 & 4.62 & 2.32 &63.9 &42.1 &23.5 &6.68 &- &- &-\\
    RAFT-Stereo~\cite{raft-stereo} &1.75 &2.89&1.96 &9.37 &4.74 &2.75 &1.27 &2.44 &0.44 &0.18\\
    CREStereo~\cite{crestereo} & 1.45 & 2.86 & 1.69 &8.25&3.71 &2.04 &1.15 &\textbf{0.98} &0.22 &\underline{0.13}\\
    EAIStereo~\cite{zhao2022eai} &1.59 &2.92 &1.81 &7.81 &3.68 &2.14 &1.09 & 2.31&1.14 &0.21\\
    DLNR~\cite{zhao2023high}   &1.60  &\underline{2.59} &1.76 &6.82 &3.20 &1.89 &1.06 &- &-&-\\
    GMStereo~\cite{xu2023unifying} &1.49 &3.14 &1.77 &23.6 &7.14 &2.96 &1.31 &1.83 &0.25 &0.19\\
    IGEV-Stereo~\cite{igevstereo} &1.38 &2.67 &1.59 &9.41 &4.83 &3.33 &2.89 &1.12 &\underline{0.21} &0.14\\
    Selective-IGEV~\cite{wang2024selective} &\underline{1.33} &2.61 &\underline{1.55} &\underline{6.53} &\underline{2.51} & \underline{1.36}&\underline{0.91} &1.23 &0.22 &\textbf{0.12}\\
    CST-Stereo (Ours) &\textbf{1.30} &\textbf{2.48} &\textbf{1.50}&\textbf{6.23}  &\textbf{2.40} &\textbf{1.27} &\textbf{0.87} &\underline{1.02} &\textbf{0.17} &\textbf{0.12}\\
    \bottomrule
    \end{tabular}
    }
    \vspace{-1mm}
    \caption{Quantitative evaluation on KITTI 2015~\cite{kitti2015}, Middlebury~\cite{middlebury} and ETH3D~\cite{eth3d} leaderboard. \textbf{Bold}: Best. \underline{Underline}: Second best.}
    \label{sota}
\vspace{-4mm}
\end{table*}

\subsection{Consistency-Aware Soft-Weighted Loss}

To take advantage of both filters and reduce the impact of inaccurate pseudo-labels, we propose consistency-aware soft-weighted loss to supervise the student model. In detail, given the weight $w_{rc}$ and $w_{ic}$ from MRPCF and IPCF, the soft filtering weight is obtained by multiplying the weights, effectively combining the strengths of both filters:
\begin{equation}
\label{eq5}
w_{soft}=w_{rc} \odot w_{ic}.
\end{equation}
The multiplication ensures that only pixels deemed reliable by both filters receive substantial weight, thus enforcing a stricter criterion for reliability. The soft filtering weights are then applied to the self-training loss function, yielding the consistency-aware soft-weighted loss: 
\begin{equation}
\label{eq6}
L_{st}=w_{soft} \odot |\hat{P}-P^O|,
\end{equation}
where $\hat{P}$ is the prediction of the student model and $P^O$ denotes the pseudo disparity map generated by the teacher model. By employing the loss, our approach leverages more unlabeled training data by incorporating less confident samples in a weighted manner, rather than discarding potentially informative examples as in hard thresholding. It not only retains more valuable data but also mitigates the risk of overfitting to highly confident yet potentially noisy labels, significantly enhancing the model's robustness and overall performance. Finally, after several optimization steps for the student, the parameters of the teacher are updated by the exponential moving average (EMA) as follows:
\begin{equation}
    \label{eq7}
    \theta_T \leftarrow \lambda \theta_T + (1-\lambda) \theta_S,
\end{equation}
where $\theta_T$ and $\theta_S$ denote weights of the teacher and student models respectively.

\begin{table*}
\centering
\scalebox{0.85}{
\begin{tabular}{l|cc|cc|cc|cc|ccc} 
\toprule
\multirow{3}{*}{Method} & \multicolumn{4}{c|}{Middlebury}  & \multicolumn{2}{c|}{\multirow{2}{*}{KITTI 2015}} & \multicolumn{2}{c|}{\multirow{2}{*}{ETH3D}} & \multicolumn{3}{c}{\multirow{2}{*}{Instereo2k}}  \\ 
\cmidrule{2-5}& \multicolumn{2}{c|}{F} & \multicolumn{2}{c|}{H} & \multicolumn{2}{c|}{}                            & \multicolumn{2}{c|}{}                       & \multicolumn{3}{c}{}  \\ 
\cmidrule{2-12}& EPE & D1               & EPE & D1               & EPE & D1                                         & EPE & D1                                    & EPE & bad 1.0 & bad 2.0  \\ 
\cmidrule{1-12}
PSMNet~\cite{psmnet}  & 40.51  & 57.93   &  9.79 &  32.18 & 4.09  & 28.64  &  2.58 &  66.34 & 70.40  &  95.95  & 93.68  \\
RAFT-Stereo~\cite{raft-stereo} & 3.84  & 15.64   & 1.44  & 11.21  & 1.13  & 5.68  & 0.27   & 2.61  &  1.01 &  11.30  & 5.82 \\
CREStereo~\cite{crestereo}  &6.59   &21.05    &1.99   &13.37   & 2.20  &7.39   &0.77   &5.10   &1.36   &17.94    &11.03  \\
EAIStereo~\cite{zhao2022eai}  & 6.16  &  18.25  & 2.15  & 11.74  & 1.87  &  6.99 & 16.30  & 23.87  & 1.42  &  17.79  & 10.89 \\
DLNR~\cite{zhao2023high}   & 6.57  &  14.46  & 1.45  & 9.46  & 3.05  & 18.91  & 9.91  &  23.00 &  0.99 &  13.20  & 6.78 \\
GMStereo~\cite{xu2023unifying} & 4.10  &  29.15  & 1.92  & 15.68  & 1.21  & 5.76  & 0.42   & 6.23  & 1.90  &  26.12  & 15.80 \\
IGEV-Stereo~\cite{igevstereo} & 5.87  & 11.85   & 1.36  & 7.21  & 1.21  & 6.03  & 0.33  &  4.06 &  1.82 & 12.11   & 5.72 \\
Selective-IGEV~\cite{wang2024selective} & 5.28  &  12.07  & 1.35  & 7.31  &  1.25 & 6.05  & 0.7  & 5.27  & 2.08  & 17.35  & 9.37 \\
\cmidrule{1-12}
CST-RAFT-Stereo  & 3.12  &  12.60  & 1.21  & 10.74  & \textbf{1.03}  & 4.93  & \textbf{0.25}  & \textbf{2.35}  & \textbf{0.60}  &  \textbf{7.61}  & \textbf{3.36} \\
CST-CREStereo  & 4.24  & 17.68   & 1.96  & 12.95  & 1.24 & \textbf{4.86}  & 0.36  & 3.68  & 1.34 &  15.08  & 8.63 \\
CST-Selective-IGEV & \textbf{1.32}  & \textbf{7.75}   & \textbf{0.77}  & \textbf{5.37}  & 1.05  & 4.91  & 0.27  & 3.54  & 0.84  &  10.16  & 5.40 \\ 
\cmidrule{1-12}
CST-RAFT-Stereo$^\dag$  & \textbf{3.78}  &  14.62  &  \textbf{1.39} & 10.59  & \textbf{1.05}  & 5.23  & \textbf{0.28}  & \textbf{2.40}  & \textbf{0.83}  &  \textbf{10.70}  & \textbf{5.28} \\
CST-CREStereo$^\dag$  & 4.81  &  20.53  &  2.01 & 13.15  & 1.10 & \textbf{5.12} & 0.30  & 3.38  & 1.08 &  16.65  & 9.25 \\
CST-Selective-IGEV$^\dag$ &  4.11 &  \textbf{9.86}  & 1.48  &  \textbf{6.47} &  1.11 & 5.21  & 0.32  & 4.12 &  0.94 & 14.62   & 6.12 \\
\bottomrule
\end{tabular}
}
\vspace{-1mm}
\caption{Generalization evaluation on KITTI 2015~\cite{kitti2015}, Middlebury~\cite{middlebury}, ETH3D~\cite{eth3d} and Instereo2k dataset. \textbf{Bold}: Best. $^\dag$: Without using any data from the target dataset. CST-RAFT, CST-CRE and CST-Selective represent CST-Stereo with RAFT-Stereo, CREStereo and Selective-IGEV as the baselines respectively.}
\label{general}
\vspace{-4mm}
\end{table*}

\section{Experiments}
\subsection{Datasets}
\textbf{Sceneflow}~\cite{dispNetC} is a synthetic dataset with over 39000 dense labeled stereo pairs. \textbf{Middlebury 2014}~\cite{middlebury} is a real-world high-resolution indoor dataset with 23 training pairs and 10 testing pairs. \textbf{KITTI 2015}~\cite{kitti2015} contains 200 training pairs and 200 testing pairs with sparse disparity maps which is collected for real-world driving scenes. \textbf{ETH3D}~\cite{eth3d} is a collection of gray-scale stereo pairs with both indoor and outdoor scenes, consisting of 27 training pairs and 20 testing pairs. \textbf{Instereo2k}~\cite{Bao2020InStereo2KAL} is a real-world indoor dataset with 2050 stereo pairs. \textbf{Holopix50k}~\cite{hua2020holopix50k} is a large-scale in-the-wild stereo dataset with nearly 50k unlabeled image pairs.

\subsection{Implementation Details}
We implement our CST-Stereo with Pytorch framework using NVIDIA A100 GPUs. Before self-training, all the models are pre-trained on the Sceneflow dataset for 200k steps with a learning rate of 2e-4. And the learning rate is set to be 1e-4 during self-training. In the training process, We use AdamW optimizer and the one-cycle learning rate schedule. For the consistency-aware soft filtering module, the $\varepsilon_1$ and $\varepsilon_2$ are set to be 5 and 10 respectively, while $\tau_1$ and $\tau_2$ are 2 and 0.5 respectively. For the EMA, the $\lambda$ is set to be 0.99 and the decay iteration number is 100. 

\subsection{Comparisons with State-of-the-art}
Following~\cite{Yue_2024_CVPR}, we evaluate the effectiveness of our method by in domain, domain adaptation, and domain generalization performance. 
\subsubsection{In Domain Performance}
\begin{figure}[t]
\centering
\includegraphics[width=0.88\columnwidth]{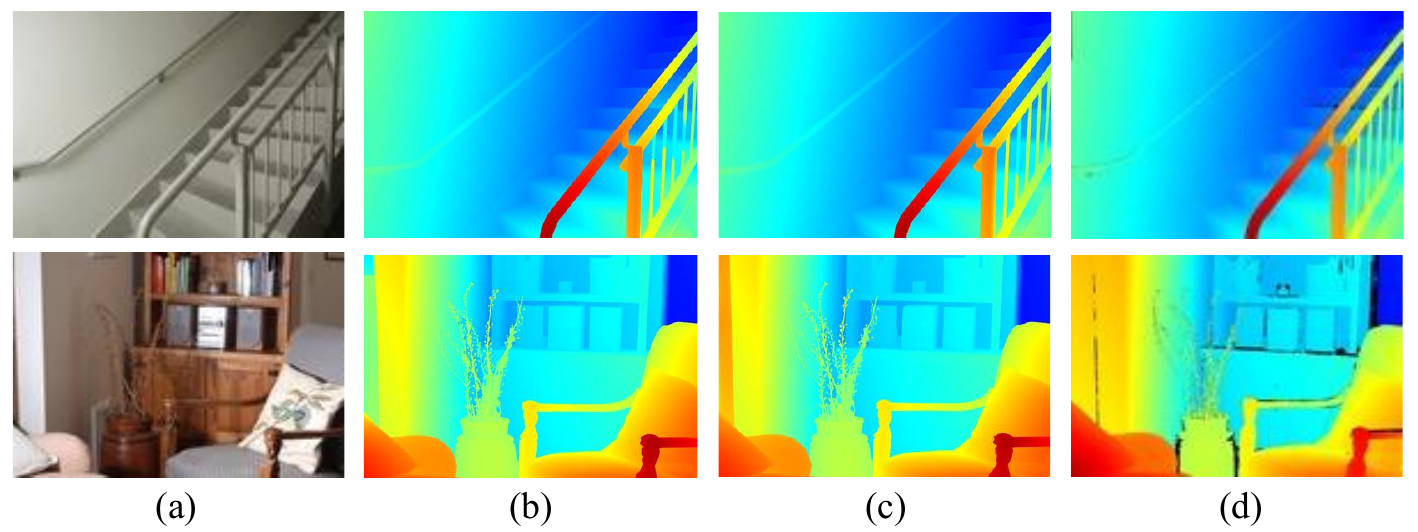} 
\vspace{-3mm}
\caption{Visualization of our method boosting the in domain performance on the Middlebury datasets. (a) Referenced images. (b) Selective-IGEV~\cite{wang2024selective}. (c) CST-Selective-IGEV. (d) Ground Truth}
\label{id}
\vspace{-4mm}
\end{figure}

Based on the current SOTA method Selective-IGEV~\cite{wang2024selective}, our method makes further improvements on in domain performance by leveraging both the labeled and unlabeled data, as shown in Tab.~\ref{sota}, with visualization results presented in Fig.~\ref{id}. We conduct extensive experiments following the same training pipeline as Selective-IGEV, which first pre-train a model on the synthetic datasets and finetune the model on a mixed dataset comprising both synthetic and real data. Differently, we also leverage unlabeled data of the real-world datasets including unannotated pixels in sparsely labeled data and some unlabeled images. 
The results are submitted to the Middlebury, KITTI2015, and ETH3D leaderboard and show evident improvements over previous SOTA methods. At the time of writing, our method ranks $3^{rd}$ on the Middlebury leaderboard. The results indicate that our approach not only matches but further enhances the capabilities of the current SOTA methods.

\subsubsection{Domain Adaptation Performance}
\begin{figure}[t]
\centering
\includegraphics[width=0.88\columnwidth]{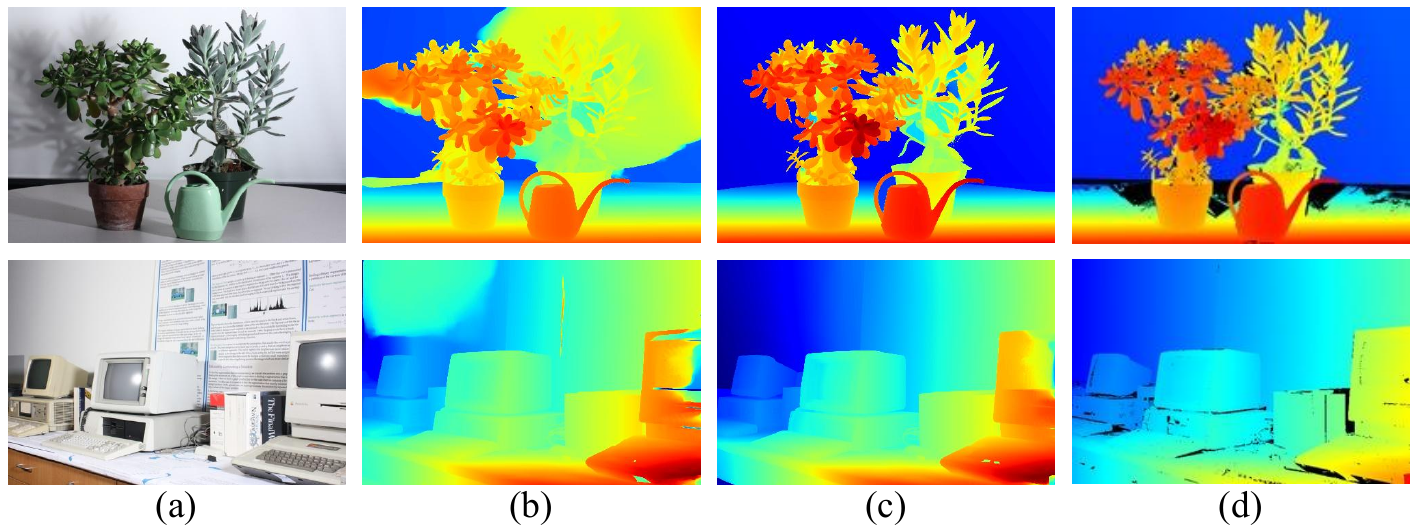} 
\vspace{-3mm}
\caption{Visualizations for the domain adaptation performance on the Middlebury datasets. (a) Referenced images. (b) Selective-IGEV~\cite{wang2024selective}. (c) CST-Selective-IGEV. (d) Ground Truth}
\label{da}
\vspace{-4mm}
\end{figure}

To further verify the effectiveness of our proposed method, we conducted extensive experiments on kinds of datasets, including Middlebury, KITTI 2015, ETH3D, and Instereo2k, to assess the domain adaptation performance of CST-Stereo. With models pre-trained on the synthetic Sceneflow dataset, our method exclusively leverages the unlabeled data from the target datasets through self-training. As illustrated in Tab.~\ref{general}, existing stereo matching methods often struggle to perform well when they do not have access to labeled data from the target datasets. Our approach significantly enhances the generalization performance of several iterative-based stereo matching methods across multiple datasets without using any labeled data from the target real-world dataset. Visualization results in Fig.~\ref{da} also demonstrate the effectiveness our method.

\subsubsection{Domain Generalization Performance}
\begin{figure}[t]
\centering
\includegraphics[width=0.88\columnwidth]{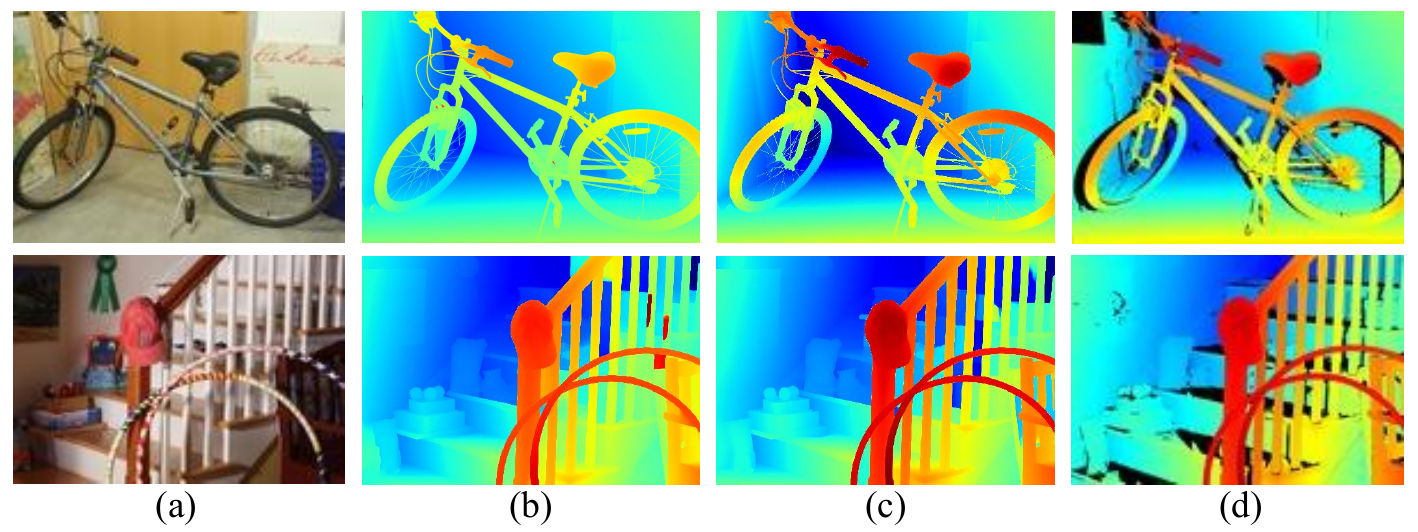} 
\vspace{-3mm}
\caption{Visualizations for the domain generalization performance on the Middlebury datasets. (a) Referenced images. (b) Selective-IGEV~\cite{wang2024selective}. (c) CST-Selective-IGEV. (d) Ground Truth}
\label{dg}
\vspace{-3mm}
\end{figure}

We also evaluate the generalization ability to unseen real-world data in Tab.~\ref{general}. Unlike the previous setups, we did not utilize any data from the target dataset, including labeled and unlabeled. Instead, we applied self-training on unlabeled data from an independent dataset, Holopix50k. Remarkably, our method still managed to improve performance on the target dataset, underscoring its effectiveness in generalizing to unseen domains without direct access to the target data. Visualization results are presented in Fig.~\ref{dg}, showing our effectiveness for domain generalization task.

\subsection{Ablation Study}
In this section, we ablate our method on the Middlebury evaluation set. All the models are pre-trained on the Sceneflow dataset, with only unlabeled data from the Middlebury dataset used in the self-training process.

\begin{table}
\centering
\small
\scalebox{0.85}{
\begin{tabular}{l|ccc|cc} 
\toprule
Model      & \begin{tabular}[c]{@{}c@{}}Self\\Training\end{tabular} & Filter & \begin{tabular}[c]{@{}c@{}}Soft \\Design\end{tabular} & EPE & D1  \\ 
\cmidrule{1-6}
Baseline   &  &  &  &  5.28  &  12.07  \\ 
\cmidrule{1-6}
ST         & $\surd$ &  &  &  6.15  & 11.76   \\
ST+Filter  & $\surd$ & $\surd$ &  & 2.85   & 9.17   \\
Full Model  & $\surd$ & $\surd$ & $\surd$ & \textbf{1.32}   & \textbf{7.75}   \\
\bottomrule
\end{tabular}
}
\vspace{-1mm}
\caption{Ablation studies for soft filtering module on Middlebury Eval dataset based on Selective-IGEV.}
\label{tab:ab1}
\vspace{-4mm}
\end{table}

\subsubsection{Ablations for consistency-aware soft filtering}
To demonstrate the effectiveness of our proposed CSF, we conduct experiments by removing the filtering step and allowing all pseudo-labels to participate in the training process, as shown in Tab.~\ref{tab:ab1}. Results show that without filtering exhibits a 48\% decrease in EPE compared to the result with filtering, and even shows a performance drop of 22\% compared to the baseline. This degradation underscores the importance of the soft filtering module. Besides, We also ablate our soft mechanism. With hard filtering, pseudo-labels are selected based on a binary decision: if the reliability score exceeds a fixed threshold, the sample is selected; otherwise, it is discarded. Conversely, our soft filtering module applies a weighted selection mechanism, where samples with higher reliability scores contribute more to the training process. Results show that soft filtering obtains further improvements compared with hard filtering, demonstrating the superiority of our soft filtering method.

\begin{table}
\centering
\scalebox{0.85}{
\begin{tabular}{l|cc} 
\toprule
Filter      & EPE & D1  \\ 
\cmidrule{1-3}
Baseline   & 5.28  &  12.07  \\ 
\cmidrule{1-3}
DDE\cite{Yue_2024_CVPR,Xu2024AUA}   &  7.47  & 15.44   \\
LRC\cite{Yue_2024_CVPR}  & 4.19   & 13.11   \\
MRPCF (Ours)  & 1.58   & 8.52   \\
IPCF (Ours)  & 1.70   & 9.18   \\
MRPCF + IPCF (Ours) & \textbf{1.32}   & \textbf{7.75}   \\
\bottomrule
\end{tabular}
}
\vspace{-1mm}
\caption{Ablation studies for filters on Middlebury Eval dataset.}
\label{tab:ab2}
\vspace{-3mm}
\end{table}

\subsubsection{Ablations for filters}
To evaluate the effectiveness of the proposed multi-resolution prediction consistency filter (MRPCF) and iterative prediction consistency filter (IPCF), we ablate the two filters respectively based on Selective-IGEV. As shown in Tab.~\ref{tab:ab2}, the inclusion of each filter independently leads to noticeable improvements in model performance, demonstrating their effectiveness. Furthermore, when both filters are combined, the model achieves the best performance, indicating that the two filters complement each other to enhance the self-training process. Besides, we compared our filtering mechanisms with those proposed in previous works~\cite{Yue_2024_CVPR,Xu2024AUA}. LRC is a filter used to remove occluded regions, and DDE is a filter that measures confidence based on the probability distribution of the cost volume. The results reveal that the previous filtering techniques are less effective for iterative-based methods, highlighting the necessity of our tailored filtering strategies for optimal performance.

\begin{table}
\centering
\scalebox{0.85}{
\begin{tabular}{l|ccccc} 
\toprule
Decay & 1 & 10 & 100 & 1000 & No decay  \\ 
\cmidrule{1-6}
EPE   & 4.15  &  1.51  & \textbf{1.32}    &  2.88    &    3.46       \\
D1    & 9.69  &  7.93  & \textbf{7.75}    &  9.13    &    10.54       \\
\bottomrule
\end{tabular}
}
\vspace{-1mm}
\caption{Ablation studies for EMA on Middlebury Eval dataset.}
\label{tab:ab3}
\vspace{-4mm}
\end{table}

\subsubsection{Ablations for EMA}
We also ablate the impact of EMA based on Selective-IGEV as shown in Tab.~\ref{tab:ab3}. The results indicate that incorporating EMA significantly improves performance. Besides, as we varied the number of delay steps for parameter updates, we observed an interesting trend: the performance initially improves as the delay increases, but eventually declines when the delay becomes too large. This phenomenon can be attributed to the stability of gradient updates being transferred to the teacher model. With too few delay steps, the updated weights may become unstable, leading to suboptimal teacher updates. Conversely, with too large a decay, the teacher's improvement becomes sluggish, failing to keep pace with the student model's learning.

\begin{table}
\centering
\scalebox{0.85}{
\begin{tabular}{c|cc|cc} 
\toprule
\multirow{2}{*}{Method} & \multicolumn{4}{c}{Middlebury}                  \\ 
\cline{2-5}
                        & \multicolumn{2}{c|}{F} & \multicolumn{2}{c}{H}  \\ 
\hline
Selective-IGEV          & 5.28 &       12.07              & 1.35 &  7.31                   \\
\cmidrule{1-5}
Ada-Selective-IGEV      & 3.65 &      11.44               & 1.56 &   6.59                  \\
ZOLE-Selective-IGEV     & 5.65 &    11.87                 & 1.16 & 6.76                    \\
CST-Selective-IGEV      & \textbf{1.32} &       \textbf{7.75}              & \textbf{0.77} &  \textbf{5.37}    \\
\cmidrule{1-5}
Ada-Selective-IGEV$^\dag$      & 4.20 &      12.58               & 1.84 &      7.51               \\
ZOLE-Selective-IGEV$^\dag$     & 6.11 &    13.05                 & \textbf{1.30} & 6.98                    \\
CST-Selective-IGEV$^\dag$      & \textbf{4.11} &       \textbf{9.86}       & 1.48 &  \textbf{6.47}                    \\
\bottomrule
\end{tabular}
}
\vspace{-1mm}
\caption{Comparision with several cross domain methods in domain adaptaion and domain generalization on Middlebury dataset.}
\label{tab:dis}
\vspace{-4mm}
\end{table}

\subsection{Discussion}
Although some unsupervised cross domain methods also enhances the model performance on out-of-domain data, we emphasize that our approach fundamentally differs from existing domain adaptation and domain generalization methods. First, while many of these methods primarily aim to enhance out-of-domain performance, they often provide limited insights into improvements in in-domain performance. In contrast, we propose a more universal strategy for leveraging unlabeled data that not only enhances out-of-domain capabilities—including domain adaptation and domain generalization—but also effectively boosts in-domain performance through the strategic use of unlabeled samples. Second, the relationship between these methods and ours is orthogonal rather than mutually exclusive. This orthogonality allows for the potential integration and exploration of our self-training approach alongside existing techniques, thereby creating opportunities for more comprehensive frameworks that can fully capitalize on unlabeled data in various contexts.
Furthermore, we demonstrate a comparison between our method and some unsupervised domain adaptation techniques in Tab.~\ref{tab:dis} The results indicate that our approach exhibits more effective improvements, especially in terms of domain generalization.
\section{Conclusion}
In this paper, we propose CST-Stereo, a novel self-training method for iterative-based stereo matching models by effectively leveraging the unlabeled data. In detail, we introduce a consistency-aware soft filtering module that can dynamically adjust the reliability of pseudo-labels to mitigate the risks associated with incorrect labels. Extensive experimental results validate the efficacy of our method, showcasing substantial improvements over current SOTA methods on multiple benchmark datasets.
Although our proposed method effectively mitigates the negative impact of incorrect pseudo-labels, the performance of the model after self-training remains constrained by the original model's capabilities and architecture. In the future, we will focus on developing more robust and adaptive model architectures. 

\section*{Acknowledgments}
This work is supported by National Key Research and Development Program of China (No. 2022ZD0160101), Shanghai Natural Science Foundation (No. 23ZR1402900), Shanghai Municipal Science and Technology Major Project (No.2021SHZDZX0103). The computations in this research were performed using the CFFF platform of Fudan University.
{
    \small
    \bibliographystyle{ieeenat_fullname}
    \bibliography{main}
}
\clearpage
\appendix
\maketitlesupplementary

\section*{Outline for Supplementary Material}
\begin{outline}
\0 Due to the page limitation for the submission paper, we present additional explanations, details and visualization results from the following aspects:
\1 ~\ref{sec:1}: More Explanations for Filtering Methods
\2 ~\ref{sec:1.1}: Preliminaries
\2 ~\ref{sec:1.2}: Explanations for Filtering Methods
\1~\ref{sec:2}: More Visualizations
\1~\ref{sec:3}: More Implementation Details
\2 ~\ref{sec:3.1}: Strong Data Augmentation
\2 ~\ref{sec:3.1}: Finetune Strategies
\end{outline}

\section{More Explanations for Filtering Methods}\label{sec:1}
\subsection{Preliminaries}\label{sec:1.1}
In general, the cost volume is a 4D tensor commonly used in stereo matching to measure the similarity between the left and right views, and it is also applied in optical flow estimation. Typically, the inputs for constructing a cost volume are the feature maps of the left and right views, with dimensions of $C \times H \times W$, where $C$ represents the channel dimension, and H and W denote the height and width of the feature maps, respectively.  The output is the cost volume with dimensions $C \times D \times H \times W$, where $C$ can sometimes be 1, depending on the specific definition of the cost volume, and $D$ represents the maximum defined disparity. 
Consequently, the cost volume represents the matching probability between each pixel in the left view and each pixel in the right view.

\begin{figure}[h]
\centering
\includegraphics[width=0.88\columnwidth]{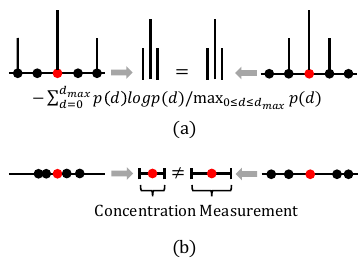} 
\caption{Comparisons between our proposed filtering method and previous methods. (a) Previous methods. (b) Our method.}
\label{sup}
\end{figure}

\subsection{Explanations for Filtering Methods}\label{sec:1.2}
Although this computational model bears some formal resemblance to classification tasks, it is inherently a regression task. This is because there exists a positional relationship between pixels along the epipolar line, meaning they are not independent of each other.  For instance, if the image width is N, a pixel calculates its matching probability with the first to the Nth pixels along the epipolar line.  In a classification task, the reliability between the first and Nth pixels is considered equivalent to that between the first and second pixels. However, in a regression task, the latter is more reliable.

Previous methods~\cite{Yue_2024_CVPR,Xu2024AUA} calculated reliability using classification approaches, treating each pixel along the epipolar line as a separate label and measuring the distribution of matching probabilities to determine reliability. However, this approach overlooks the spatial positional relationship between pixels along the epipolar line, treating all pixels as equivalent labels. As illustrated in Fig.~\ref{sup} (a), the reliability of the two distributions on the left and right is inconsistent, yet prior methods would consider them identical.  Our method, on the other hand, utilizes multiple predicted values to determine the spatial concentration of the predicted values along the epipolar line, providing a more reasonable measure of prediction reliability, as shown in Fig.~\ref{sup} (b).

\begin{figure*}[t]
\centering
\includegraphics[width=0.88\linewidth]{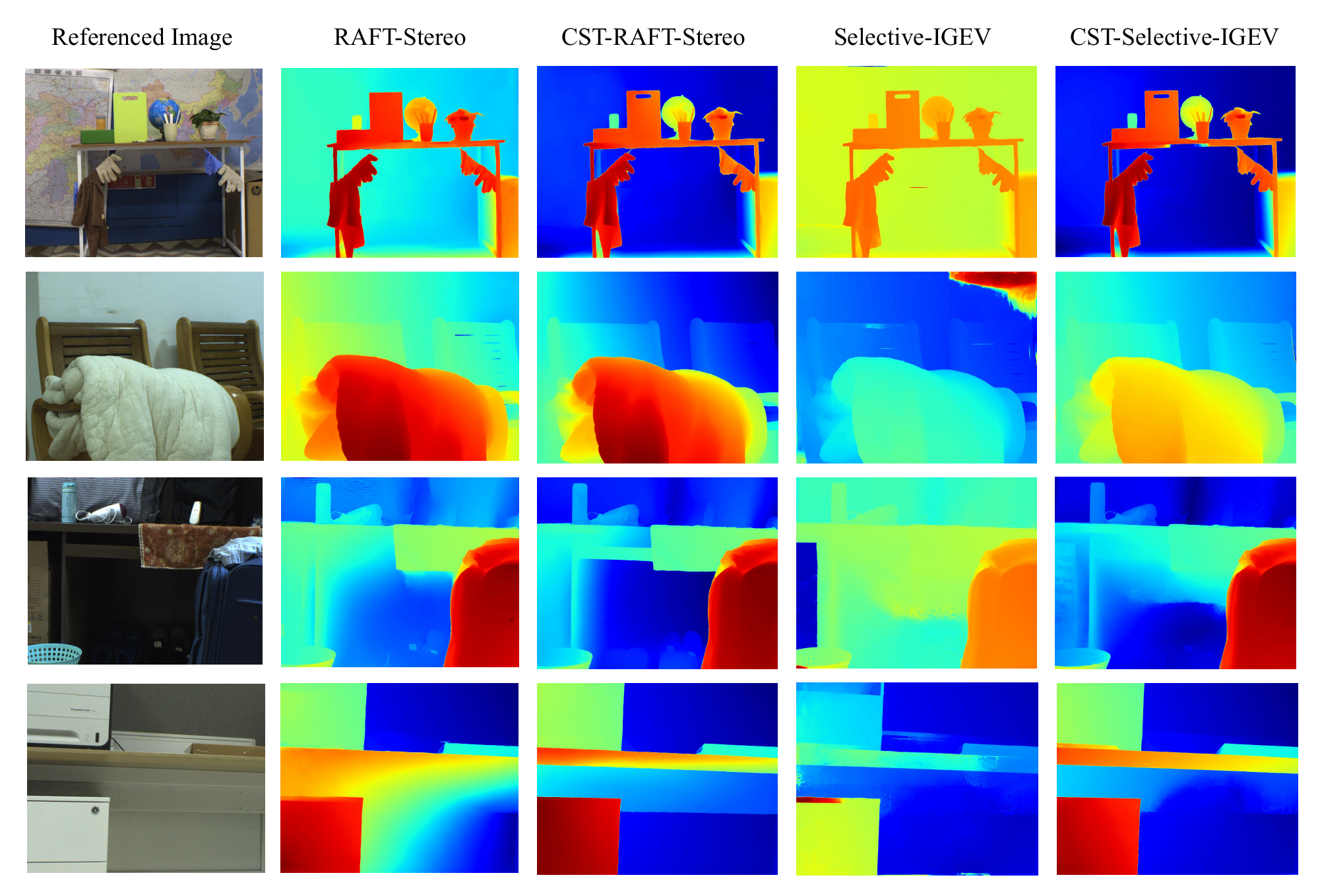} 
\caption{Cross-domain generalization visualization results on the Instereo2k dataset.}
\label{visual}
\end{figure*}


\section{More Visualizations}\label{sec:2}
To further validate the effectiveness of our CST-Stereo, we show more cross-domain generalization visualization results on the InStereo2K dataset~\cite{Bao2020InStereo2KAL}, as shown in Fig.~\ref{visual}. Our consistency-aware self-training is applied on RAFT-Stereo~\cite{raft-stereo} and Selective-IGEV~\cite{wang2024selective} respectively. Before applying our self-training, both models struggled with ambiguous boundaries and often misclassified regions, leading to substantial errors. Notably, our method enhances the distinction between foreground and background, leading to more precise segmentation and a marked reduction in errors. This improvement indicates that our approach can bring a robust correction to previously problematic areas.

\section{More Implementation Details}\label{sec:3}
\subsection{Strong Data Augmentation}\label{sec:3.1}
To better decouple the outputs of the teacher and student models, we apply strong data augmentation to the inputs of the student model. Specifically, we perform the following augmentations:
1) Saturation Adjustment: We adjust the saturation of the left and right images with a range of 0 to 1.4.
2) Random Occlusion: We apply random occlusion to the right image, where the occluded region's pixels are replaced with the mean value of the region.
3) Brightness Transformation: We adjust the brightness of the left and right images within a range of 0.8 to 1.2.
4) Gaussian Noise and Blur: We randomly add Gaussian noise and apply Gaussian blur to left and right images.
These augmentations are designed to introduce variations in the student's inputs, encouraging the model to learn more robust and generalized features independently from the teacher's output.

\subsection{Finetune Strategies}\label{sec:3.2}
In our experiments to compare with state-of-the-art methods on the leaderboard of several benchmark datasets, we follow the setting of our baseline model Selective-IGEV~\cite{wang2024selective} in the finetune stage. We also leverage unlabeled data from real-world datasets, including unannotated pixels in sparsely labeled data and some unlabeled images. Specifically, there are unlabeled pixels in the Middlebury, KITTI, Instereo2k, and ETH3D datasets and unlabeled images in the Middlebury 2001, 2005, and 2006 datasets.
\subsubsection{Middlebury.}
For the Middlebury dataset, we first finetune our pre-trained model on the mixed Tartan Air~\cite{wang2020tartanair}, CREStereo Dataset~\cite{crestereo}, Scene Flow~\cite{mayer2016large}, Falling things~\cite{tremblay2018falling}, InStereo2k~\cite{Bao2020InStereo2KAL}, CARLA HR-VS~\cite{yang2019hierarchical}, and Middlebury~\cite{middlebury} datasets 200k steps using a crop size of 384 $\times$ 512 with a batch size of 8. Then we finstune the model on the mixed CREStereo Dataset, Falling Things, InStereo2k, CARLA HR-VS, and Middlebury datasets using a crop size of 384 $\times$ 768 with a batch size of 8 for another 100k steps.

\subsubsection{KITTI 2015.}
For the KITTI 2015 dataset, following Selective-Stereo~\cite{wang2024selective}, we finetune the pretrained model on the mixed dataset of KITTI 2012~\cite{kitti2012} and KITTI 2015~\cite{kitti2015} with a batch size of 8 for 50k steps.

\subsubsection{ETH3D.}
For the ETH3D dataset, following~\cite{wang2024selective}, we finetune the pre-trained model on the mixed Tartan Air~\cite{wang2020tartanair}, CREStereo Dataset~\cite{crestereo}, Scene Flow~\cite{mayer2016large}, Sintel Stereo~\cite{butler2012naturalistic}, InStereo2k~\cite{Bao2020InStereo2KAL}, and ETH3D~\cite{eth3d} datasets for 300k steps. Then we fintune it on the mixed CREStereo Dataset, InStereo2k, and ETH3D datasets for another 90k steps.


\end{document}